\documentclass[10pt,twocolumn,letterpaper]{article}

\usepackage{cvpr}
\usepackage{times}
\usepackage{epsfig}
\usepackage{graphicx}
\usepackage{amsmath}
\usepackage{amssymb}

\usepackage{subfigure}
\usepackage{breqn}
\usepackage{multirow}
\usepackage{threeparttable}
\usepackage{makecell}
\usepackage{verbatim}

\usepackage[pagebackref=true,breaklinks=true,letterpaper=true,colorlinks,bookmarks=false]{hyperref}

\cvprfinalcopy 

\def\cvprPaperID{2053} 

\begin{document}

\title{Deep Spatial Gradient and Temporal Depth Learning for Face Anti-spoofing}

\author{
Zezheng Wang$^1$~~~~Zitong Yu$^{2}$~~~~Chenxu Zhao$^{3,}$\thanks{\ denotes the corresponding author.}~~~~Xiangyu Zhu$^4$~~~~Yunxiao Qin$^{5}$~~~~Qiusheng Zhou$^6$\\
Feng Zhou$^1$~~~~Zhen Lei$^4$\\[1mm]
\normalsize{$^1$AIBEE~~~~$^2$CMVS, University of Oulu~~~~$^3$Academy of Sciences, Mininglamp Technology}\\
\normalsize{$^4$CBSR\&NLPR, CASIA~~~~$^5$Northwestern Polytechnical University~~~~$^6$JD Digits}\\[1mm]
{\tt\small \{zezhengwang, fzhou\}@aibee.com~~~~ zitong.yu@oulu.fi~~~~zhaochenxu@mininglamp.com}\\
{\tt\small\{xiangyu.zhu, zlei\}@nlpr.ia.ac.cn~~~~qyxqyx@mail.nwpu.edu.cn~~~~ zhouqiusheng3@jd.com}
}

\maketitle

\begin{abstract}
Face anti-spoofing is critical to the security of face recognition systems. Depth supervised learning has been proven as one of the most effective methods for face anti-spoofing. Despite the great success, most previous works still formulate the problem as a single-frame multi-task one by simply augmenting the loss with depth, while neglecting the detailed fine-grained information and the interplay between facial depths and moving patterns. 
In contrast, we design a new approach to detect presentation attacks from multiple frames based on two insights: 1) detailed discriminative clues (e.g., spatial gradient magnitude) between living and spoofing face may be discarded through stacked vanilla convolutions, and 2) the dynamics of 3D moving faces provide important clues in detecting the spoofing faces.
The proposed method is able to capture discriminative details via Residual Spatial Gradient Block (RSGB) and encode spatio-temporal information from Spatio-Temporal Propagation Module (STPM) efficiently. Moreover, a novel Contrastive Depth Loss is presented for more accurate depth supervision. To assess the efficacy of our method, we also collect a Double-modal Anti-spoofing Dataset (DMAD) which provides actual depth for each sample. The experiments demonstrate that the proposed approach achieves state-of-the-art results on five benchmark datasets including OULU-NPU, SiW, CASIA-MFSD, Replay-Attack, and the new DMAD. Codes will be available at \url{https://github.com/clks-wzz/FAS-SGTD}.
\end{abstract}

\section{Introduction}
\label{sec:intro}

Face recognition technology has become the most indispensable component in many interactive AI systems for their convenience and human-level accuracy. However, most of existing face recognition systems are easily to be spoofed through presentation attacks (PAs) ranging from printing a face on paper (print attack) to replaying a face on a digital device (replay attack) or bringing a 3D-mask (3D-mask attack). Therefore, not only the research community but also the industry has recognized face anti-spoofing
~\cite{Frischholz2000BiolD, frischholz2003avoiding,li2004live_videolet,liu20163d,nowara2017ppgsecure,Chetty2006Multi,hu2018exploring, casiasurf,agarwal2016face, li2004live,ReplayAttack,wen2015face,Qin2020MetaFas,yu2020auto,GuoZXLWL19}
as a critical role in securing the face recognition system.

\begin{figure}[t]
 \centering
  \includegraphics[width=8.5cm,height=2.7cm]{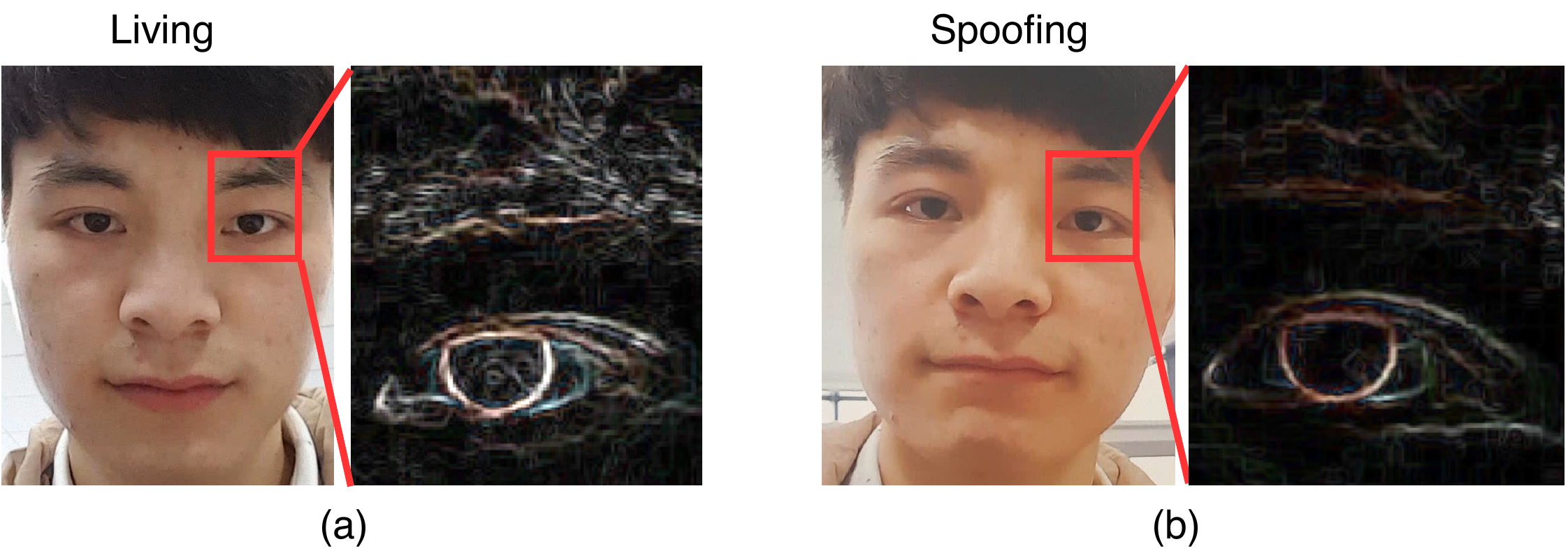}
  \caption{Spatial gradient magnitude difference between living (a) and spoofing (b) face. Notice that the large difference in gradient maps despite their similarities in the original RGB images.}
  \label{fig:IntroGradient}
\end{figure}

In the past few years, both traditional methods
~\cite{Pereira2012LBP,Patel2016Secure,Boulkenafet2017Face_SURF}
and CNN-based methods
~\cite{Lucena2017Transfer,Nagpal2018A,Gan20173D,liu2019AuroraGuard,song2019discriminative}
have shown effectiveness in discriminating between the living and spoofing face. They often formalize face anti-spoofing as a binary classification between spoofing and living images. 
However, these approaches are challenging to explore the nature of spoofing patterns, such as the loss of skin details, color distortion, 
moir$\rm\acute{e}$ pattern, and spoofing artifacts.

In order to overcome this issue, many auxiliary depth supervised face anti-spoofing methods have been developed. Intuitively, the images of living faces contain face-like depth, whereas the images of spoofing faces in print and by replaying carriers only have planar depth. 
Thus, Atoum \etal~\cite{Atoum2018Face} and Liu \etal ~\cite{Liu2018Learning} propose single-frame depth supervised CNN architectures, and improve the presentation attack detection (PAD) accuracy.

\begin{figure}[t]
  \includegraphics[width=0.48\textwidth]{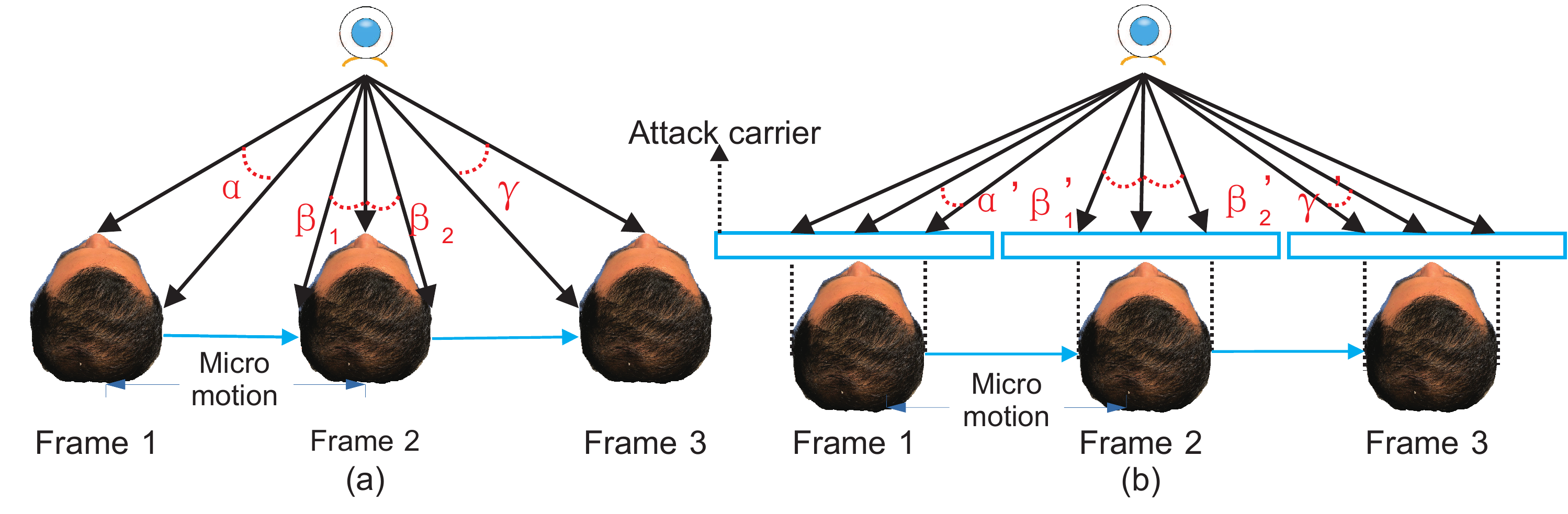}
  \caption{
  Temporal depth difference between live and spoof (print attack here) scenes. The change in camera viewpoint can result in facial motion among different keypoints. In the living scene (a), the angle $\alpha$ between nose and right ear is getting smaller,
  while the angle ${\beta}_1$ between left ear and nose is getting larger.
  However, in the spoofing scene (b), the observation could be different $\alpha'< {\beta}'_2$, and ${\beta}'_1 > {\gamma}'$. 
  }
  \label{fig:intro}
\end{figure}

By surveying the past face anti-spoofing methods, we notice there are two problems that have not yet been fully solved:
\textbf{1)} Traditional methods usually design local descriptors for solving PAD while modern deep learning methods can learn to extract relatively high-level semantic features instead. Despite their effectiveness, we argue that low-level fine-grained patterns can also play a vital role in distinguishing living and spoofing faces, e.g. the spatial gradient magnitude shown in Fig.~\ref{fig:IntroGradient}. So how to aggregate local fine-grained information into convolutional networks is still unexplored for face anti-spoofing task.
\textbf{2)} Recent depth supervised face anti-spoofing methods~\cite{Atoum2018Face,Liu2018Learning} estimate facial depth based on a single frame and leverage depth as dense pixel-wise supervision in a direct manner. We argue that the {\em virtual} discrimination of depth between living and spoofing faces can be explored more adequately by multiple frames. A vivid and exaggerated example with assumed micro motion is illustrated in Fig.~\ref{fig:intro}. 

To address the problems, we present a novel depth supervised spatio-temporal network with Residual Spatial Gradient Block (RSGB) and Spatio-Temporal Propagation Module (STPM). Inspired by ResNet~\cite{he2016deep} , our RSGB aggregates learnable convolutional features with spatial gradient magnitude via shortcut connection. 
As a result, both local fine-grained patterns and traditional convolution features can be captured via stacked RSGB.
To better utilize the information from multiple frames, STPM is designed for propagating short-term and long-term spatio-temporal features into depth reconstruction. To supervise the models with facial depth more effectively, we propose a Contrastive Depth Loss (CDL) to learn the topography of facial points.

We believe that the accuracy of facial depth directly affects the establishment of the relationship between temporal motion and facial depth. So we collect a double-modal anti-spoofing dataset named Double-modal Anti-spoofing Dataset (DMAD) which provides actual depth map for each sample. Extensive experiments are conducted to show that actual depth is more appropriate for monocular PAD than the generated depth. Note that this paper mainly focuses on the planar attack, which is the most common in practice.

We summarize the main contributions below.
\begin{itemize}
\setlength{\itemsep}{0pt}
\setlength{\parsep}{0pt}
\setlength{\parskip}{0pt}
\item We propose a novel depth supervised architecture to capture discriminative details via Residual Spatial Gradient Block (RSGB) and encode spatio-temporal information from Spatio-Temporal Propagation Module (STPM) efficiently from monocular frame sequences.  
\item We develop a Contrastive Depth Loss to learn the topography of facial points for depth supervised PAD.
\item We collect a double-modal dataset to verify that the actual depth is more appropriate for monocular PAD than the generated depth. This indicates an insight that collecting corresponding depth image to the RGB image brings benefit to the progress of the monocular PAD.
\item We demonstrate the state-of-the-art performance by our method on widely used face anti-spoofing benchmarks.
\end{itemize}

\section{Related Work}
\label{sec:related}
Roughly speaking, previous face anti-spoofing works generally fall into three categories: binary supervised, depth supervised, and temporal-based methods.

\noindent \textbf{Binary supervised Methods}\quad Since face anti-spoofing is essentially a binary classification problem, most of previous anti-spoofing methods train a classifier under binary supervision, e.g., spoofing face as 0 and living face as 1. 
The early works usually rely on hand-crafted features, such as LBP~\cite{Pereira2012LBP,Pereira2013Can,Maatta2011Face}, 
SIFT~\cite{Patel2016Secure},
SURF~\cite{Boulkenafet2017Face_SURF}, HoG~\cite{Komulainen2014Context,Yang2013Face}, 
DoG~\cite{Peixoto2011Face,Tan2010Face}, 
and traditional classifiers, such as SVM and Random Forests. Because of the sensitiveness of manually-engineered features, traditional methods often generalize poorly across varied conditions such as camera devices, lighting conditions and presentation attack instruments (PAIs). 
Recently, CNN has emerged as a powerful tool in face anti-spoofing tasks with the help of hardware advancement and data abundance. 
For instance, in early works like \cite{Li2017An,Patel2016Cross}, pre-trained VGG-face model is fine-tuned to extract features in a binary-classification setting.
However, most of them consider face anti-spoofing as a binary classification problem with cross-entropy loss, which easily learns the arbitrary patterns such as screen bezel.

\begin{figure*}[!htb]
  \centering
  \includegraphics[width=17.6cm,height=5.2cm]{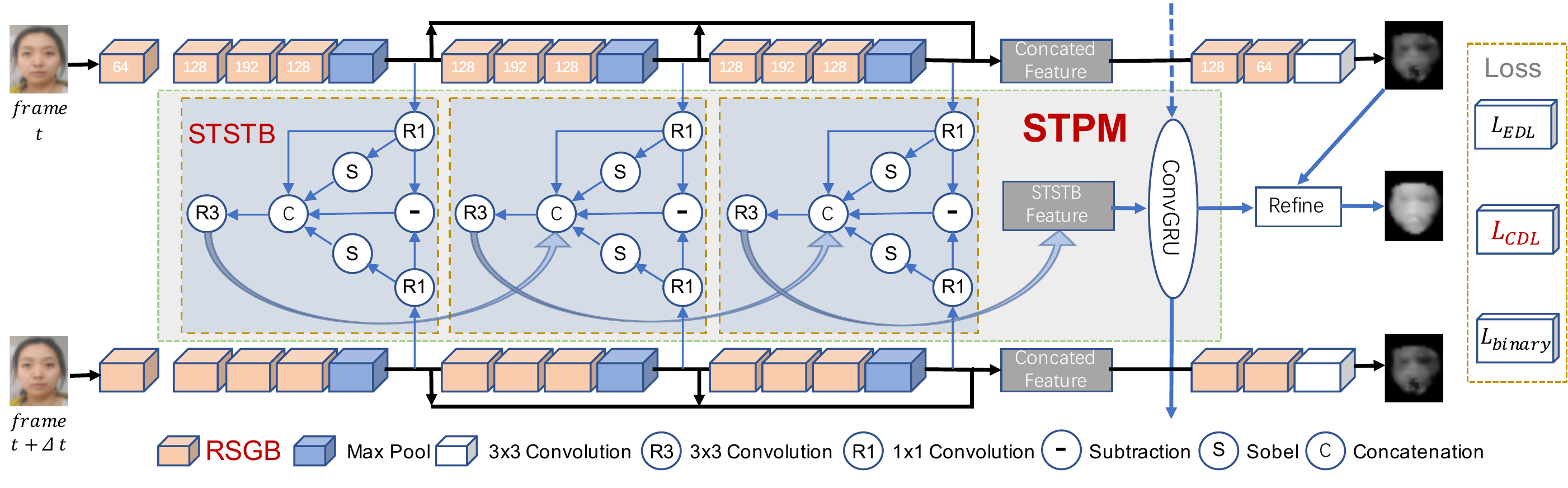}
  \caption{Illustration of the overall framework. The inputs are consecutive frames with a fixed interval. Each frame is processed by cascaded RSGB with a shared backbone which generates a corresponding coarse depth map. The number in RSGB cubes denotes the output channel number of RSGB. STPM is plugged between frames for estimating the temporal depth, which is used for refining the corresponding coarse depth map. The framework works well by learning with the overall loss functions.}
  \label{fig:pipeline}
\end{figure*}

\noindent \textbf{Depth supervised Methods} \quad  Compared with the binary setting, depth supervised methods aim to learn more faithful patterns. In \cite{Atoum2018Face}, the depth map of a face is utilized as a supervisory signal for the first time. They propose a two-stream CNN-based approach for face anti-spoofing, by extracting both the patch features and holistic depth maps from the face images. It shows that depth estimation is beneficial for modeling face anti-spoofing to obtain promising results, especially on higher-resolution images. 
In another work~\cite{Liu2018Learning}, the authors propose a face anti-spoofing method by augmenting spatial facial depth as an auxiliary supervision along with temporal rPPG signals. 
More recently, \cite{jourabloo2018face} attempts to learn spoof noise and depth for generalized face anti-spoofing. 
However, these methods take stacked vanilla convolutional networks as the backbone and fail to capture the rich detailed patterns for depth estimation.

\noindent \textbf{Temporal-based Methods}  \quad  Temporal information plays a vital role in face anti-spoofing tasks. Most of the prior works focus on the movement of key parts of the face. 
For example in ~\cite{Pan2007Eyeblink,Patel2016Cross}, the eye-blinking fact is used to predict spoofing. 
However, these methods are vulnerable to replay attacks since they heavily rely on some heuristic assumptions about the nature of these attacks. 
More general approaches like 3D convolution \cite{Gan20173D} or LSTM \cite{Xu2016Learning,yang2019face} have recently been used to distinguish the live from spoof images.
In addition, optical flow magnitude map and Shearlet feature have been taken as inputs in \cite{Feng2016Integration} to the CNN due to the obvious difference in flow patterns between living and spoofing faces. Based on the different color changes between the living and spoofing face videos, rPPG~\cite{li2016generalized,Liu2018Learning,lin2019face} features are also explored for PAD.
To the best of our knowledge, no depth supervised temporal-based methods has ever been proposed for face anti-spoofing task.

\section{The Proposed Approach}
In this section, we first present our advanced depth-supervised spatio-temporal network structure, including Residual Spatial Gradient Block (RSGB) and Spatio-Temporal Propagation Module (STPM). Then our proposed novel Contrastive Depth Loss (CDL) and the overall loss would be demonstrated.
\subsection{Network Structure}
Designed in an end-to-end depth supervised fashion, our proposed framework takes $N_f$-frame face images as input and predicts the corresponding depth map directly. As shown in Fig.~\ref{fig:pipeline}, the backbone is composed of cascaded RSGB followed by pooling layers, intending to extract fine-grained spatial features in low-level, mid-level and high-level, respectively. Then these multi-level features are concatenated to predict coarse depth map for each frame. 

In order to capture rich dynamic information, STPM is plugged between frames. Short-term Spatio-Temporal Block (STSTB) picks up spatio-temporal features from adjacent frames while ConvGRU propagates these short-term features in a multi-frame long-term view. Finally, the temporal depth maps estimated from STPM are used to refine the coarse depth from the backbone.

\begin{figure}[t]
 \centering
  \includegraphics[width=7.8cm,height=2.2cm]{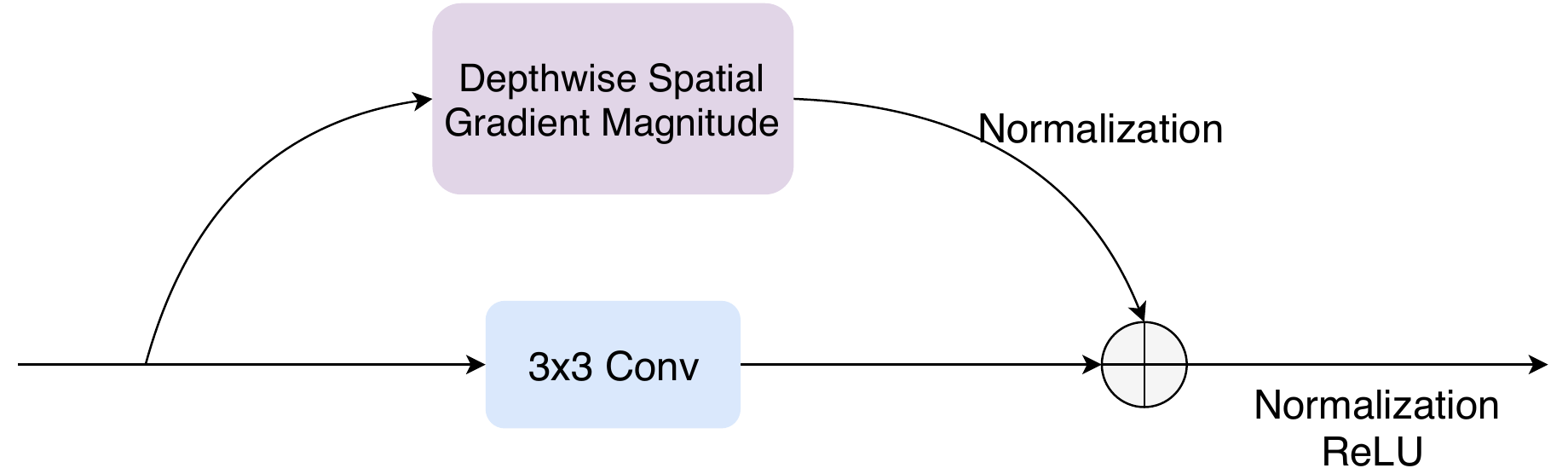}
  \caption{Residual spatial gradient block.}
  \label{fig:Residual}
\end{figure}

\subsubsection{Residual Spatial Gradient Block}

Fine-grained spatial details are vital for distinguishing the bona fide and attack presentations. As illustrated in Fig.~\ref{fig:IntroGradient}, the gradient magnitude response between the living (Fig.~\ref{fig:IntroGradient}(a)) and spoofing (Fig.~\ref{fig:IntroGradient}(b)) face is quite different, which gives the insight to design a residual spatial gradient block (RSGB) for capturing such discriminative clues. 
In this paper, we take the well-known Sobel \cite{kanopoulos1988design} operation to compute gradient magnitude. 
In a nutshell, the horizontal and vertical gradients can be derived from the following convolutions respectively:
\begin{small}
\begin{equation}
F_{hor}(x) \! = \! \begin{bmatrix}
 -1& 0 &+1 \\ 
 -2& 0 &+2\\ 
 -1& 0 & +1
\end{bmatrix} \! \odot x,F_{ver}(x) \! = \! \begin{bmatrix}
 -1& -2 &-1 \\ 
 0& 0 &0\\ 
 +1& +2 & +1
\end{bmatrix} \! \odot x,
\label{eq:sobel}
\end{equation}
\end{small}
where $\odot$ denotes the depthwise convolution operation, and $x$ represents the input feature maps. As shown in Fig.~\ref{fig:Residual}, our RSGB adopts the advanced shortcut connection structure to aggregate the learnable convolutional features with gradient magnitude information, which intends to enhance representation ability of fine-grained spatial details. It can be formulated as
\begin{equation}
y= \phi ( \mathcal{N} ( F(x,\left \{ W_{i} \right\})+ \mathcal{N} ({F_{hor}(x^{'})^2+F_{ver}(x^{'})^2}) ) ),
\label{eq:residual}
\end{equation}
where $x$ represents the input features maps while $x^{'}$ denotes the feature maps altered through 1x1 convolution, which intends to keep the consistent channel numbers for subsequent residual addition. $y$ denotes the output feature maps. $\mathcal{N}$ and $\phi$ denote the normalization and Relu layer, respectively. The function $F(x,\left \{ W_{i} \right\})$ represents the residual gradient magnitude mapping to be learned. Note that the proposed RSGB is able to plug in both image and feature levels, extracting rich spatial context for depth regression task.

\subsubsection{Spatio-Temporal Propagation Module} Virtual discrimination of depth between living and spoofing faces can be explored adequately by multiple frames. Therefore, we design STPM to extract multi-frame spatio-temporal features for depth estimation, via Short-term Spatio-Temporal Block (STSTB) and ConvGRU.

\textbf{STSTB.} \quad
As illustrated in Fig.~\ref{fig:pipeline}, STSTB extracts the generalized short-term spatio-temporal information by fusing five kinds of features: the current compressed features $F_l(t)$, the current spatial gradient features $F_l^S(t)$, the future spatial gradient features $F_l^S(t+\triangle t)$, the temporal gradient features $F_l^{T}(t)$, and the STSTB features from the previous level $STSTB_{l-1}(t)$.
The fused features can provide weighted spatial and temporal information in a learnable/adaptive way. In this paper, the spatial and temporal gradients are implemented with Sobel-based depthwise convolution (similar to Eq.~\ref{eq:sobel}) and element-wise subtraction of temporal features, respectively. Note that the 1x1 convolutions intend to compress the channel number with more efficiency.

Different from the related OFF~\cite{Sun2018Optical} work, we consider both spatial gradient of the current compressed features $F_l^S(t)$ and future spatial gradient features $F_l^S(t+\triangle t)$ while OFF only considers $F_l^S(t)$. Moreover, current compressed feature $F_l(t)$ itself also plays an important role in recovering the fine depth map, which is concatenated in STSTB as well. The detailed comparison between STSTB and OFF will be studied in Sec.~\ref{sec:Experimental}, which shows the advancement of STSTB especially for depth-supervised face anti-spoofing task.

\textbf{ConvGRU.} \quad
As short-term information between two consecutive frames from STSTB has limited representation ability, it is natural to use the recurrent neural network to capture long-range spatio-temporal context. However, the classical LSTM and GRU~\cite{chung2014empirical} neglect the spatial information in hidden units. 
In consideration of the spatial neighbor relationship in the hidden layers, ConvGRU is conducted for propagating the long-range spatio-temporal information. 
ConvGRU can be described as below:
\begin{align}
\centering
&R_t = \sigma(K_r \otimes [H_{t-1}, X_t]),  \ \nonumber
U_t = \sigma(K_u \otimes [H_{t-1}, X_t]),\nonumber\\
&\hat{H}_t = \tanh(K_{\hat{h}} \otimes [R_t * H_{t-1}, X_t]),\nonumber\\
&H_t = (1 - U_t)*H_{t-1} + U_t * \hat{H}_t,
\label{eq:ConvGRU}
\end{align}
where $X_t, H_t, U_t$ and $R_t$ are the matrix of input, output, update gate and reset gate, $K_r, K_u, K_{\hat{h}}$ are the kernels in the convolution layer, $\otimes$ is convolution operation, $*$ denotes element wise product, and $\sigma$ denotes the sigmoid activation function.

\subsubsection{Depth Map Refinement} 
Forwarding the RSGB based backbone and STPM for a given $N_f$-frame input, we could obtain the corresponding coarse depth maps $\textbf{\rm{D}}_{single}^{t}$ and temporal depth maps $\textbf{\rm{D}}_{multi}^{t}$, respectively, where $t \in [1,N_f-1]$ denotes the $t$-th frame. Then $\textbf{\rm{D}}_{multi}^{t}$ is utilized to refine $\textbf{\rm{D}}_{single}^{t}$ in a weighted summation manner: 
\begin{equation}
\textbf{\rm{D}}_{refined}^{t} = (1-\alpha) \cdot \textbf{\rm{D}}_{single}^{t} + \alpha \cdot \textbf{\rm{D}}_{multi}^{t},
\alpha \in [0,1],
\label{eq:multi_ratio}
\end{equation}
where $\alpha$ is the trade-off weight between $\textbf{\rm{D}}_{single}^{t}$ and $\textbf{\rm{D}}_{multi}^{t}$. The higher value of $\alpha$ indicates the more importance about the multi-frame spatio-temporal features. Finally, $N_f-1$ refined depth maps $\{\textbf{\rm{D}}_{refined}^{t}\}^{N_f-1}_{t=1}$ are obtained.

\subsection{Loss Function}
Besides designing the network architecture, we also need an appropriate loss function to guide the network training. One major step-forward of the current study is that we design a novel Contrastive Detph Loss, which is able to combine with classical loss, further boosting performance.

\subsubsection{Contrastive Detph Loss}
\begin{figure}[t]
  \centering
\includegraphics[width=8.8cm,height=2.8cm]{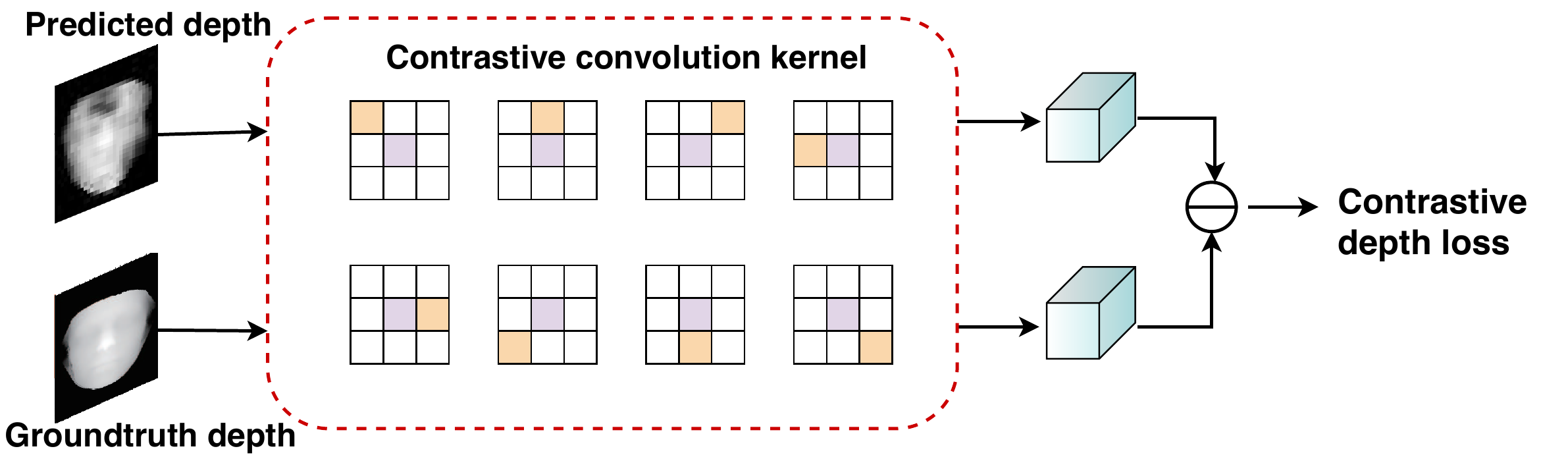}
  \caption{Contrastive Depth Loss. The purple, yellow, and white pieces indicate 1, -1, and 0, respectively. There are totally eight contrastive convolution kernels in CDL.}
  \label{fig:K_contrast}
\end{figure}

In the classical depth-based face anti-spoofing, Euclidean Distance Loss (EDL) is usually used for pixel-wise supervision, which is formulated:
\begin{equation}
\begin{split}
{L}_{EDL} = & ||\textbf{\rm{D}}_{P} - \textbf{\rm{D}}_{G}||_{2}^{2},
\label{eq:euclidean_distance_loss}
\end{split}
\end{equation}
where $\textbf{\rm{D}}_{P}$ and $\textbf{\rm{D}}_{G}$ are the predicted depth and groundtruth depth, respectively. EDL applies supervision on the predicted depth based on pixel one by one, ignoring the depth difference among adjacent pixels. Intuitively, EDL merely assists the network to learn the absolute distance between the objects to the camera. However, the distance relationship of different objects is also important to be supervised for the depth learning. Therefore, as shown in Fig.~\ref{fig:K_contrast}, we propose the Contrastive Depth Loss (CDL) to offer extra strong supervision, which improves the generality of the depth-based face anti-spoofing model:
\begin{equation}
\small
\begin{split}
{{L}}_{CDL} = & \sum_i{||\textbf{\rm{K}}_{i}^{CDL} \odot \textbf{\rm{D}}_{P} - \textbf{\rm{K}}_{i}^{CDL} \odot \textbf{\rm{D}}_{G}||_{2}^{2}},
\label{eq:contrast_depth_loss}
\end{split}
\end{equation}
where $\textbf{\rm{K}}_{i}^{CDL}$ is the ${ith}$ contrastive convolution kernel, $i \in [0,7]$. The details of the kernels can be found in Fig.~\ref{fig:K_contrast}.

\subsubsection{Overall Loss}
In view of the potentially unclear depth map, we hereby consider a binary loss when looking for the difference between living and spoofing depth map. Note that the depth supervision is decisive, whereas the binary supervision takes an assistant role to discriminate the different kinds of depth maps.
\begin{equation}
{{L}}_{binary} = - \textbf{{\rm{B}}}_{G} * log(fcs( \textbf{{\rm{D}}}_{avg}  )),
\label{eq:binary_loss}
\end{equation}
\begin{equation}
{{{L}}_{overall}} = \beta \cdot {{{L}}_{binary}} + (1-\beta) \cdot ({{L}}_{EDL}+{{L}}_{CDL}),
\label{eq:mutil_loss}
\end{equation}
where $\textbf{{\rm{B}}}_{G}$ is the binary groundtruth label, $\textbf{{\rm{D}}}_{avg}$ is the pool averaged map of $\{\textbf{\rm{D}}_{refined}^{t}\}^{N_f-1}_{t=1}$, and $fcs$ denotes two fully connected layers and one softmax layer after the element-wise averaged depth maps, which outputs the logits of two classes, $\beta$ is the hyper-parameter to trade-off binary loss and depth loss in the final overall loss ${{L}_{overall}}$.

\section{Double-modal Anti-spoofing Dataset}

\begin{figure}[t]
  \includegraphics[width=8.4cm,height=4.0cm]{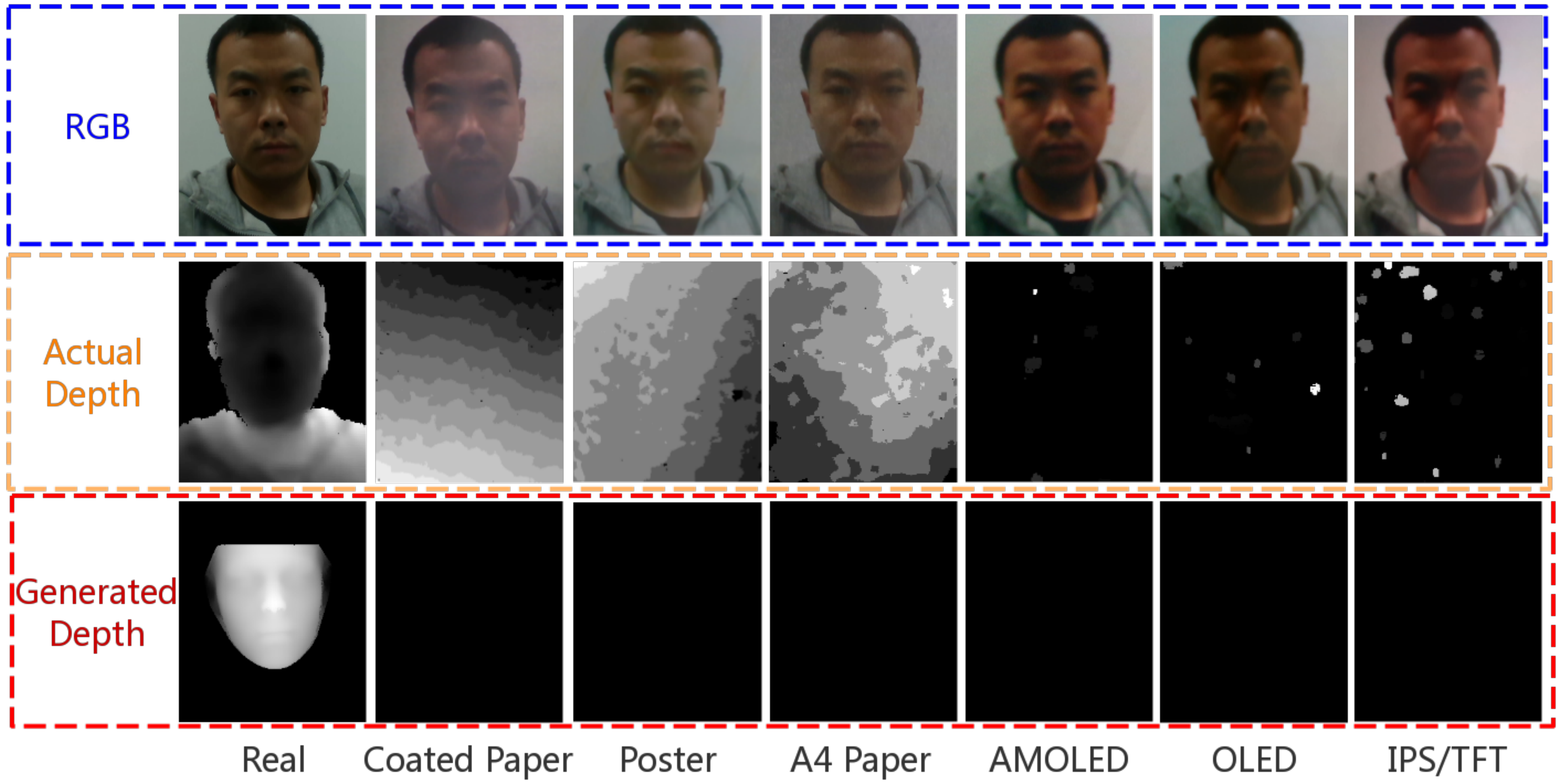}
  
  \caption{Some examples of DMAD. The actual depth is more precise than the generated depth.
  }
  \label{fig:DfAD}
\end{figure}

In this work, we collect a real double-modal dataset (RGB and Depth). 
There are three kinds of display materials in replay attack: 
AMOLED screen, OLED screen, IPS/TFT screen. 
Meanwhile, three kinds of paper materials in print attacks are adopted: high-quality A4 paper, coated paper, and poster paper. The capture camera is RealSense SR300, which can offer corresponding RGB and Depth images. 
There are 300 subjects, each of which is recorded in three sessions and contains one real category and two attack categories (print and replay). 
Totally, we obtain 2700 samples (4\textasciitilde 12 seconds videos) in less than two months with two human workers. Tab.~\ref{tab:DFAD_detail} demonstrates the details of DMAD, and Fig.~\ref{fig:DfAD} shows some corresponding examples.

\begin{table}
\caption{The details of our collected DMAD. This protocol of splitting subsets aims to evaluate the generalization of methods under unseen presentation materials. } 
\resizebox{0.50\textwidth}{!}{
\begin{tabular}{|c|c|c|c|c|c|}
\hline
Subset & Subject & Session &Modal Types & Presentation Material & \# of live/attack vid. \\
\hline
\multirow{2}{*}{Train} &1\textasciitilde 100  &1\textasciitilde 3 &RGB, Depth  &A4 Paper, AMOLED &900 \\
       &101\textasciitilde 200 &1\textasciitilde 3  &RGB, Depth  &Coated Paper, OLED &900 \\
\hline
Test &201\textasciitilde 300  &1\textasciitilde 3  &RGB, Depth  &Poster Paper, IPS/TFT &900 \\
\hline
\end{tabular}
}
\label{tab:DfAD_detail}
\end{table}

\section{Experiments}
\subsection{Databases and Metrics}
\subsubsection{Databases}
Five databases - OULU-NPU~\cite{Boulkenafet2017OULU,Boulkenafet2017OULU_compare}, SiW~\cite{Liu2018Learning}, CASIA-MFSD~\cite{Zhang2012A}, Replay-Attack~\cite{ReplayAttack}, DMAD are used in our experiment. OULU-NPU~\cite{Boulkenafet2017OULU} is a high-resolution database, consisting of 4950 real access and spoofing videos and containing four protocols to validate the generalization of models. SiW~\cite{Liu2018Learning} contains more live subjects and three protocols are used for testing. CASIA-MFSD~\cite{Zhang2012A} and Replay-Attack~\cite{ReplayAttack} both contain low-resolution videos. 

\subsubsection{Performance Metrics}
In OULU-NPU and SiW dataset, we follow the original protocols and metrics for a fair comparison. OULU-NPU, SiW and DMAD utilize 1) Attack Presentation
Classification Error Rate $APCER$, which evaluates the highest error among all PAIs (e.g.
print or display), 2) Bona Fide Presentation Classification Error Rate $BPCER$, which evaluates the error of real access data, and 3) $ACER$~\cite{ACER}, which evaluates the mean of $APCER$ and $BPCER$:
\begin{equation}
ACER=\frac{APCER+BPCER}{2}.
\end{equation}

HTER is adopted in the cross-database testing between CASIA-MFSD and Replay-Attack, evaluating the mean of False Rejection Rate (FRR) and False Acceptance Rate (FAR):
\begin{equation}
HTER=\frac{FRR+FAR}{2}.
\end{equation}

\subsection{Implementation Details}

\subsubsection{Depth Generation}
Dense face alignment method PRNet~\cite{Feng2018Joint} is adopted to estimate the 3D shape of the living face and generate the facial depth map $\textbf{\rm{D}}_G \in \mathbb{R}^{32 \times 32}$. A typical sample can be found in the third row of Fig.~\ref{fig:DfAD}. To distinguish living faces from spoofing faces, at the training stage, we normalize living depth map in a range of $[0, 1]$, while setting spoofing depth map to 0, which is similar to ~\cite{Liu2018Learning}.

\subsubsection{Training Strategy}
The proposed method is trained with a two-stage strategy: \emph{Stage 1:} We train the backbone with cascaded RSGB by the depth loss $L_{EDL}$ and $L_{CDL}$, in order to learn a fundamental representation to predict coarse depth maps. \emph{Stage 2:} We fix the parameters of the backbone, and train the STPM part by the overall loss $L_{overall}$ for refining depth maps. Our networks are fed by $N_f$ frames, which are sampled by an interval of three frames. This sampling interval makes sampled frames maintain enough temporal information in the limited GPU memory.

\subsubsection{Testing Strategy}
For the final classification score, we feed the sequential frames into the network and obtain depth maps ${\{\textbf{\rm{D}}_{refined}^t\}}^{N_f-1}_{t=1}$ and the living logits $\hat{b}$ in $fcs(\textbf{\rm{D}}_{avg})$. The final living score can be obtained by:
\begin{equation}
score = \beta \cdot \hat{b} + (1-\beta) \cdot
\frac{ \sum_{t=1}^{N_f-1}{||\textbf{\rm{D}}_{refined}^t * \textbf{\rm{M}}^t||_1} }{ N_f-1 },
\label{eq:testing_inference}
\end{equation}
where $\beta$ is the same as that in equation~\ref{eq:mutil_loss}, $\textbf{\rm{M}}^t$ is the mask of face at frame $t$ , which can be generated by the dense face landmarks in PRNet~\cite{Feng2018Joint}, and the second module denotes that we compute the mean of depth values in the facial areas as one part of the score.
\subsubsection{Hyper-parameter Setting}
Our proposed method is implemented in Tensorflow, with a learning rate of 1e-4 for single-frame part and 1e-2 for multi-frame part. The batch size of single-frame part is 48, and that of multi-frame part is 2 with $N_f$ being 5 in our experiment. Adadelta optimizer is used in our training procedure, with $\rho$ as 0.95 and $\epsilon$ as 1e-8. We set $\alpha=0.6$ and $\beta=0.8$ by our experimental experience.

\subsection{Experimental Comparison}
\label{sec:Experimental}
\subsubsection{Ablation Study}
\begin{figure}[t]
  \includegraphics[width=7.8cm,height=3.2cm]{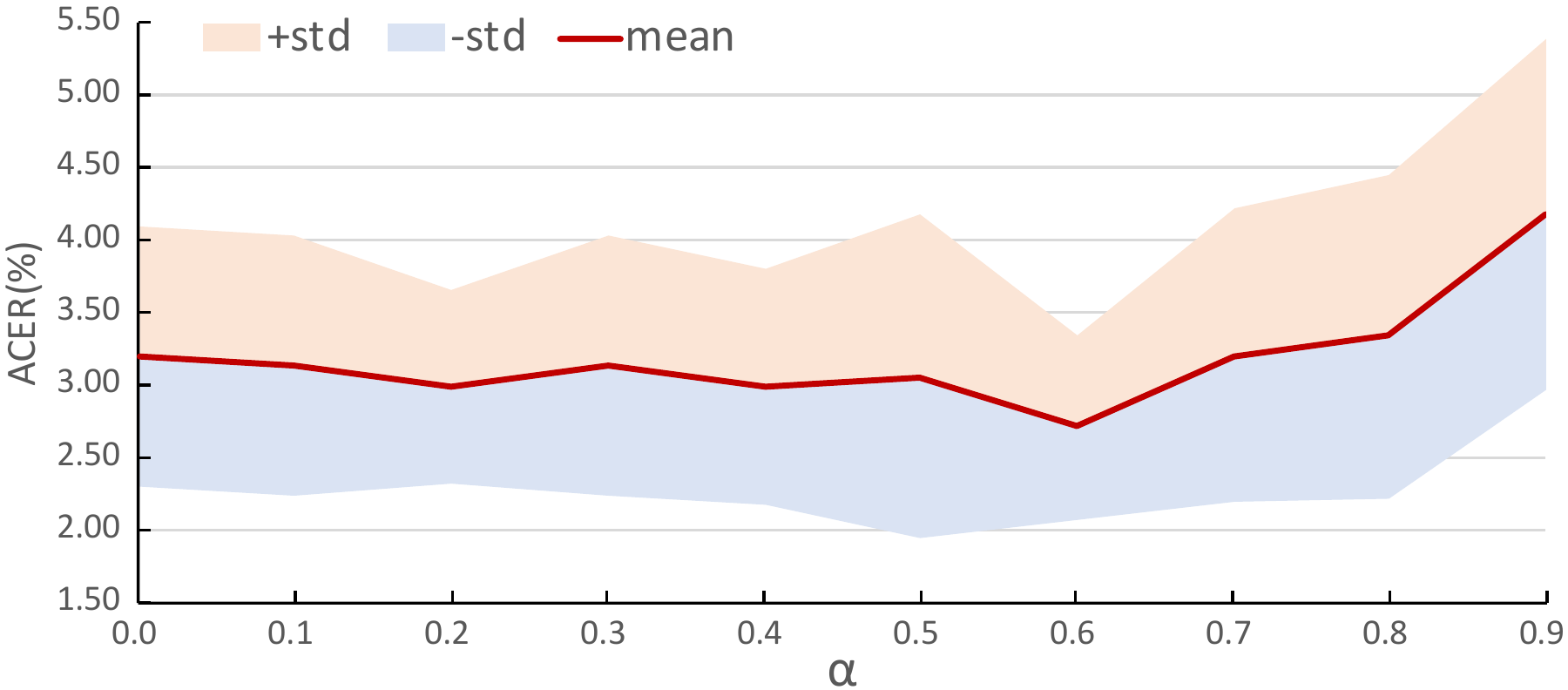}
  \caption{Ablation study of $\alpha$ in Eq.~\ref{eq:multi_ratio} on OULU-NPU Protocol 3. The red line denotes the mean ACER(\%) value while the orange/blue area denotes the range of standard deviation.
  }
  \label{fig:multi-ratio}
\end{figure}

\begin{table}[t]
\centering
\caption{The results of ablation study on OULU-NPU Protocol 3.}
\resizebox{0.48\textwidth}{!}{
\begin{tabular}{|l|c c c c c c|c|}
\hline
Module &$L_{CDL}$ &RSGB &STSTB &OFF &ConvGRU &$L_{binary}$ &ACER(\%)\\
\hline
Model 1 &\   &\   &\  &\  &\  &\  &6.25$\pm$3.20 \\
Model 2 &$\surd$ &\ &\  &\  &\  &\  &5.07$\pm$1.83 \\
Model 3 &$\surd$ &$\surd$ &\  &\  &\  &\ &3.19$\pm$0.90 \\
Model 4 &$\surd$ &$\surd$ &$\surd$ &\  &\  &\  &2.99$\pm$0.72 \\
Model 5 &$\surd$ &$\surd$ &$\surd$ &\   &$\surd$  &\  &2.85$\pm$0.49 \\
Model 6 &$\surd$ &$\surd$ &\ &$\surd$  &$\surd$  &$\surd$  &3.20$\pm$1.00 \\
Model 7 &$\surd$ &$\surd$ &$\surd$ &\  &$\surd$ &$\surd$  &\textbf{2.71$\pm$0.63} \\
\hline
\end{tabular}
}
\label{tab:ablation}
\end{table}

Seven architectures are implemented to demonstrate the efficacy of vital parts (i.e., RSGB, STPM and loss functions) in the proposed method. As shown in Tab.~\ref{tab:ablation}, Model 1 can be treated as a raw baseline, consisting of a backbone network with stacked vanilla convolutions. Model 2 is supervised with extra contrastive depth loss. Based on Model 2, vanilla convolutions are replaced by RSGB in Model 3. Moreover, Model 4 and Model 5 are designed for validating the effectiveness of STSTB and ConvGRU. In Model 6, STSTB is replaced by normal OFF~\cite{Sun2018Optical}. Model 7 is our complete architecture with all modules and losses.

\textbf{Efficacy of the Modules and Loss Functions.} \quad It can be seen from Tab.~\ref{tab:ablation} that Model 2 outperforms Model 1, which means our proposed CDL helps to estimate more accurate depth maps. With the progressive lower ACER of Model 3, Model 4 and Model 5, it is clear that RSGB, STSTB and ConvGRU contribute to extract effective discriminative features respectively. Finally, in comparison between Model 5 and Model 7, binary supervision indeed assists to distinguish live vs. spoof.

\textbf{STSTB vs. OFF.} \quad As illustrated in Tab.~\ref{tab:ablation}, Model 7 with STSTB surpasses Model 6 with OFF for a large margin, which implies that the current and future gradient information is valuable for spatio-temporal face anti-spoofing task. Model 6 even achieves inferior result compared with Model 3, indicating that it is challenging to design an effective temporal module for depth regression task.

\textbf{Importance of Spatio-temporal Information for Depth Refinement.} \quad It can be seen from Eq.~\ref{eq:multi_ratio} that the depth map refinement is conducted in a weighted summation manner and hyperparameter $\alpha$ controls the contribution of the temporal depth maps predicted by STPM. 
As shown in Fig.~\ref{fig:multi-ratio}, with appropriate valuel of $\alpha$, the model can be benefited from spatio-temporal information and achieves better performance than that using only spatial information ($\alpha = 0.0$). And the best performance can be obtained when $\alpha = 0.6$.

\textbf{Influence of Sampling Interval in Spatio-temporal Architecture.} We conduct experiments on one sub-protocol of Protocol 3 with various sampling intervals ($\triangle t$). When $\triangle t$ euqals to $1, 3, 5,$ and $ 7$ frame(s), the ACER is 3.347\%, 2.927\%, 4.223\%, and 2.934\%, respectively.
The ACER is the lowest when $\triangle t=3$, which is used as the default setting for the following intra- and cross-database testing.

\subsubsection{Intra-database Testing}
\label{sec:intra_testing}
\begin{table}[t]
\caption{The results of intra-database testing on four protocols of OULU-NPU. For a fair comparison, here the results STASN~\cite{yang2019face} trained without extra private dataset are reported. }
\resizebox{0.45\textwidth}{!}{
\begin{tabular}{|c|c|c|c|c|}
\hline
Prot. & Method & APCER(\%) & BPCER(\%) & ACER(\%) \\
\hline
\multirow{6}{*}{1} &CPqD~\cite{boulkenafet2017competition} &2.9 &10.8 & 6.9 \\
        &GRADIANT~\cite{boulkenafet2017competition} &1.3 &12.5 & 6.9 \\
        &STASN~\cite{yang2019face} &1.2 &2.5 & 1.9 \\
        &Auxiliary~\cite{Liu2018Learning} &1.6 &1.6 & 1.6 \\
        &FaceDs~\cite{jourabloo2018face} &1.2 &1.7 & 1.5 \\
        &\textbf{OURs} &2.0 &0.0 & \textbf{1.0} \\
\hline
\multirow{6}{*}{2} &MixedFASNet~\cite{boulkenafet2017competition} &9.7 &2.5 & 6.1 \\
       &FaceDs~\cite{jourabloo2018face} &4.2 &4.4 & 4.3 \\
       &Auxiliary~\cite{Liu2018Learning} &2.7 &2.7 & 2.7 \\
       &GRADIANT~\cite{boulkenafet2017competition} &3.1 &1.9 & 2.5 \\
       &STASN~\cite{yang2019face} &4.2 &0.3 & 2.2 \\
       &\textbf{OURs} &2.5 &1.3 & \textbf{1.9} \\
\hline
\multirow{4}{*}{3} &MixedFASNet~\cite{boulkenafet2017competition} &5.3$\pm$6.7 &7.8$\pm$5.5 &6.5$\pm$4.6 \\
       &GRADIANT~\cite{boulkenafet2017competition} &2.6$\pm$3.9 &5.0$\pm$5.3 &3.8$\pm$2.4 \\
       &FaceDs~\cite{jourabloo2018face} &4.0$\pm$1.8 &3.8$\pm$1.2 &3.6$\pm$1.6 \\
       &Auxiliary~\cite{Liu2018Learning} &2.7$\pm$1.3 &3.1$\pm$1.7 &{2.9}$\pm$1.5 \\
       &STASN~\cite{yang2019face} &4.7$\pm$3.9 &0.9$\pm$1.2  &2.8$\pm$1.6 \\
       &\textbf{OURs} &3.2$\pm$2.0 &2.2$\pm$1.4 &\textbf{2.7}$\pm$\textbf{0.6} \\
\hline
\multirow{4}{*}{4} &Massy\_HNU~\cite{boulkenafet2017competition} &35.8$\pm$35.3 &8.3$\pm$4.1 &22.1$\pm$17.6 \\
       &GRADIANT~\cite{boulkenafet2017competition} &5.0$\pm$4.5 &15.0$\pm$7.1 &10.0$\pm$5.0 \\
       &Auxiliary~\cite{Liu2018Learning} &9.3$\pm$5.6 &10.4$\pm$6.0 &9.5$\pm$6.0 \\
       &STASN~\cite{yang2019face} &6.7$\pm$10.6 &8.3$\pm$8.4  &7.5$\pm$4.7 \\
       &FaceDs~\cite{jourabloo2018face} &1.2$\pm$6.3 &6.1$\pm$5.1 &5.6$\pm$5.7 \\
       &\textbf{OURs} &6.7$\pm$7.5 &3.3$\pm$4.1 &\textbf{5.0$\pm$2.2} \\
\hline
\end{tabular}
}
\label{tab:OULU}
\end{table}

\begin{table}
\centering
\caption{The results of intra-database testing on three protocols of SiW~\cite{Liu2018Learning}. } 
\resizebox{0.45\textwidth}{!}{
\begin{tabular}{|c|c|c|c|c|}
\hline
Prot. & Method & APCER(\%) & BPCER(\%) & ACER(\%) \\
\hline
\multirow{3}{*}{1} &Auxiliary~\cite{Liu2018Learning} &3.58 &3.58 &3.58 \\
       &STASN~\cite{yang2019face} &-- &-- &1.00 \\
       &\textbf{OURs} &0.64 &0.17 &\textbf{0.40} \\
\hline
\multirow{3}{*}{2} &Auxiliary~\cite{Liu2018Learning} &0.57$\pm$0.69 &0.57$\pm$0.69 &0.57$\pm$0.69 \\
       &STASN~\cite{yang2019face} &-- &-- &0.28$\pm$0.05 \\
       &\textbf{OURs} &0.00$\pm$0.00 &0.04$\pm$0.08 &\textbf{0.02$\pm$0.04} \\
\hline
\multirow{3}{*}{3} &STASN~\cite{yang2019face} &-- &-- &12.10$\pm$1.50 \\
       &Auxiliary~\cite{Liu2018Learning} &8.31$\pm$3.81 &8.31$\pm$3.80 &8.31$\pm$3.81 \\
       &\textbf{OURs} &2.63$\pm$3.72 &2.92$\pm$3.42 &\textbf{2.78$\pm$3.57} \\
\hline
\end{tabular}
}
\label{tab:SiW}
\end{table}



We compare the performance of intra-database testing on OULU-NPU, SiW and DMAD datasets. There are four protocols in OULU-NPU for evaluating the generalization of the developed face presentation attack detection (PAD) methods. Protocol 1 and Protocol 2 are designed to evaluate the generalization of PAD methods under previously unseen illumination scene and under unseen attack medium (e.g., unseen printers or displays), respectively. Protocol 3 utilizes a Leave One Camera Out (LOCO) protocol, in order to study the effect of the input camera variation. Protocol 4 considers all the above factors and integrates all the constraints from protocols 1 to 3, so protocol 4 is the most challenging.

\textbf{Results on OULU-NPU.} \quad  As shown in Tab.~\ref{tab:OULU}, our proposed method ranks first on all 4 protocols, which indicates the proposed method performs well at the generalization of the external environment, attack mediums and input camera variation. It's worth noting that our model has the lowest mean and std of ACER in protocol 3 and 4, indicating its good accuracy and stablity.

\textbf{Results on SiW.} \quad   Tab.~\ref{tab:SiW} compares the performance of our method with two state-of-the-art methods Auxiliary~\cite{Liu2018Learning} and STASN~\cite{yang2019face} on SiW dataset. According to the purposes of three protocols on SiW and the results in Tab.~\ref{tab:SiW}, we can see that our method performs great advantages on the generalization of (a) variations of face pose and expression, (b) variations of different spoof mediums, (c) cross presentation attack instruments.

\textbf{Results on DMAD.} \quad   The results of intra-database testing on DMAD are shown in Tab.~\ref{tab:DfAD}. In this experiment, we still set spoofing depth map to zero when training the actual depth model. 
Tab.~\ref{tab:DfAD} shows that the ACER(\%) of multi-frame model (Model 7) supervised by actual depth obtains 1.78 lower than that supervised by generated depth. This demonstrates the actual depth map brings benefit to the improvement of monocular face anti-spoofing. 

\begin{table}
\centering
\caption{The results of intra-database testing on DMAD. } 
\resizebox{0.45\textwidth}{!}{
\begin{tabular}{|c|c|c|c|c|}
\hline
Method & Depth Map & APCER(\%) & BPCER(\%) & ACER(\%) \\
\hline
\multirow{2}{*}{Model 7} &Generated &9.17 &3.48 &6.33 \\
       &Actual &6.36 &2.75 &\textbf{4.55} \\
\hline
\end{tabular}
}
\label{tab:DfAD}
\end{table}

\newcommand{\tabincell}[2]{\begin{tabular}{@{}#1@{}}#2\end{tabular}}
\begin{table}
\caption{The results of cross-database testing between CASIA-MFSD and Replay-Attack. The evaluation metric is HTER(\%).}
\resizebox{0.46\textwidth}{!}{
\begin{tabular}{|c|c|c|c|c|}
\hline
\multirow{2}{*}{Method} &Train &Test &Train &Test\\
\cline{2-3} \cline{4-5} &\tabincell{c}{CASIA-\\MFSD} &\tabincell{c}{Replay-\\Attack} &\tabincell{c}{Replay-\\Attack} &\tabincell{c}{CASIA-\\MFSD}\\
\hline
Motion~\cite{Pereira2013Can}
&\multicolumn{2}{c|}{50.2} &\multicolumn{2}{c|}{47.9} \\
LBP-1~\cite{Pereira2013Can}
&\multicolumn{2}{c|}{55.9} &\multicolumn{2}{c|}{57.6} \\
LBP-TOP~\cite{Pereira2013Can}
&\multicolumn{2}{c|}{49.7} &\multicolumn{2}{c|}{60.6} \\
Motion-Mag~\cite{bharadwaj2013computationally}
&\multicolumn{2}{c|}{50.1} &\multicolumn{2}{c|}{47.0} \\
Spectral cubes~\cite{pinto2015face}
&\multicolumn{2}{c|}{34.4} &\multicolumn{2}{c|}{50.0} \\
CNN~\cite{Yang2014Learn}
&\multicolumn{2}{c|}{48.5} &\multicolumn{2}{c|}{45.5} \\
LBP-2~\cite{boulkenafet2015face}
&\multicolumn{2}{c|}{47.0} &\multicolumn{2}{c|}{39.6} \\
STASN~\cite{yang2019face}
&\multicolumn{2}{c|}{31.5} &\multicolumn{2}{c|}{30.9} \\
Colour Texture~\cite{Boulkenafet2017Face}
&\multicolumn{2}{c|}{30.3} &\multicolumn{2}{c|}{37.7} \\
FaceDs~\cite{jourabloo2018face}
&\multicolumn{2}{c|}{28.5} &\multicolumn{2}{c|}{41.1} \\
Auxiliary~\cite{Liu2018Learning}
&\multicolumn{2}{c|}{27.6} &\multicolumn{2}{c|}{28.4} \\
\textbf{OURs}
&\multicolumn{2}{c|}{\textbf{17.0}} &\multicolumn{2}{c|}{\textbf{22.8}} \\
\hline
\end{tabular}
}
\label{tab:cross-testing}
\end{table}

\begin{table}[t]
\centering
\caption{The results of cross-database testing from SiW to OULU-NPU dataset. } 
\resizebox{0.42\textwidth}{!}{
\begin{tabular}{|c|c|c|c|c|}
\hline
Prot. & Method & APCER(\%) & BPCER(\%) & ACER(\%) \\
\hline
\multirow{2}{*}{1} &Auxiliary~\cite{Liu2018Learning} &-- &-- &10.0 \\
       &\textbf{OURs} &1.7 &13.3 &\textbf{7.5} \\
\hline
\multirow{2}{*}{2} &Auxiliary~\cite{Liu2018Learning} &-- &-- &14.1 \\
       &\textbf{OURs} &9.7 &14.2 &\textbf{11.9} \\
\hline
\multirow{2}{*}{3} &Auxiliary~\cite{Liu2018Learning} &-- &-- &\textbf{13.8$\pm$5.7} \\
       &\textbf{OURs} &17.5$\pm$4.6 &11.7$\pm$12.0 &{14.6$\pm$4.8} \\
\hline
\multirow{2}{*}{4} &Auxiliary~\cite{Liu2018Learning} &-- &-- &10.0$\pm$8.8 \\
       &\textbf{OURs} &0.8$\pm$1.9 &10.0$\pm$11.6 &\textbf{5.4$\pm$5.7} \\
\hline
\end{tabular}
}
\label{tab:SiWOulu}
\end{table}

\subsubsection{Cross-database Testing}

We utilize four datasets (CASIA-MFSD, Replay-Attack, SiW and OULU-NPU) to perform cross-database testing for measuring the generalization potential of the models. 

\textbf{Results on CASIA-MFSD and Replay-Attack.} \quad   In this experiment, there are two cross-database testing protocols. One is training on the CASIA-MFSD and testing on Replay-Attack, which we name protocol CR; the other is training on the Replay-Attack and testing on CASIA-MFSD, which we name protocol RC. In Tab.~\ref{tab:cross-testing}, it is shown that HTER(\%) of our proposed method is 17.0 on protocol CR and 22.8 on protocol RC, reducing 38.4\% and 19.7\% respectively compared with the previous state of the art. The improvement of performance on cross-database testing demonstrates the good generalization of proposed method.

\textbf{Results from SiW to OULU-NPU.} \quad  Here, It is shown that the cross-database testing results trained on SiW and tested on OULU-NPU in Tab.~\ref{tab:SiWOulu}. It can be seen that our method outperforms Auxiliary~\cite{Liu2018Learning} on three protocols (decrease 2.5\%, 2.2\% and 4.6\% of ACER on protocol 1, protocol 2 and protocol 4, respectively). In protocol 3, ACER of our method is 14.6$\pm$4.8\% and slightly higher than that of Auxiliary.
Considering the rPPG used in Auxiliary method, it may also be a good choice combined with proposed method.

\subsubsection{Visualization and Analysis}
\begin{figure}[t]
  \centering
  \includegraphics[width=7.6cm,height=3.8cm]{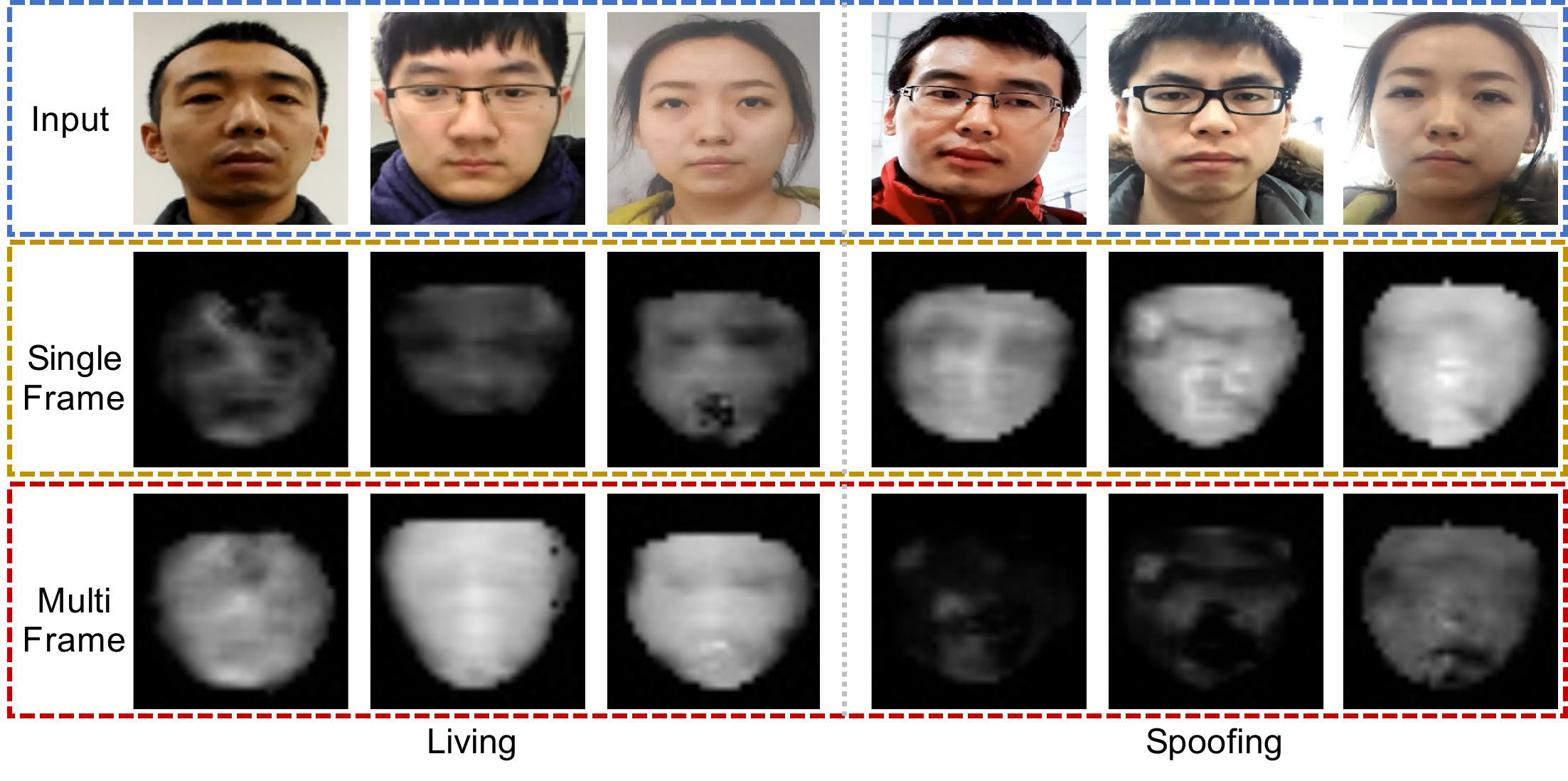}
  \caption{The generated results of hard samples in OULU-NPU. The predicted coarse depth maps from stacked RSGB backbone and temporal depth maps from STPM are illustrated in the second and third row, respectively. }
  \label{fig:result_maps}
\end{figure}

\begin{figure}[!htbp]
  \centering
  \includegraphics[width=8.4cm,height=3.2cm]{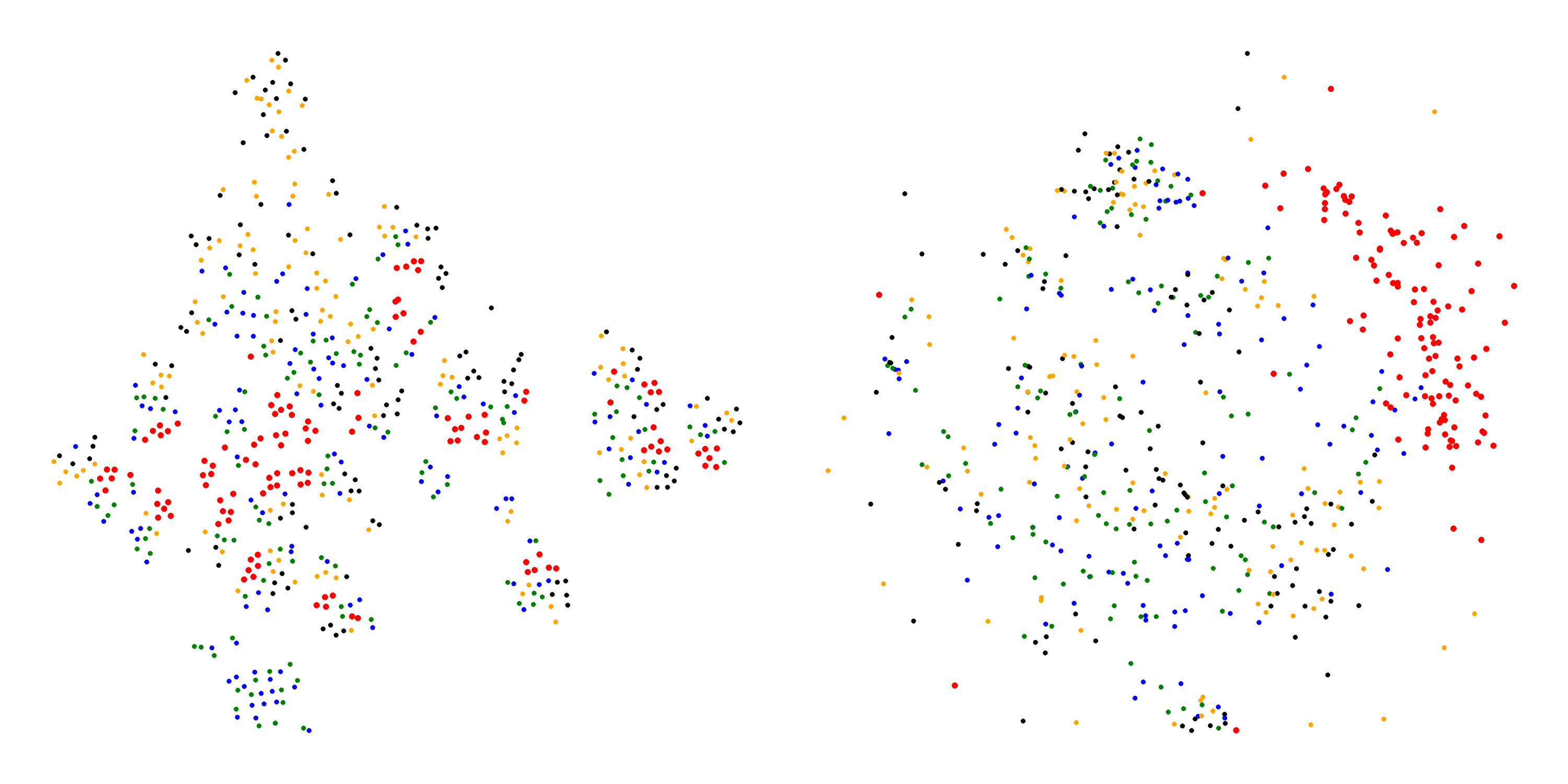}
  \caption{Feature distribution visualization of the testing videos on OULU-NPU Protocol 1 using t-SNE~\cite{maaten2008visualizing}
  . Left: features w/o RSGB, Right: features w/ RSGB. Color indicates $red$=live, $green$=printer1, $blue$=printer2, $orange$=display1, $black$=display2.}
  \label{fig:scatter_plot}
\end{figure}

The predicted depth maps of hard samples in OULU-NPU Protocol 3 are partly visualized in Fig.~\ref{fig:result_maps}. It can be seen that some samples are difficult for the single-frame PAD to be detected. In contrary, our multi-frame methods with STPM can estimate more precise depth maps than those of single-frame method. 
The difference of depth images from real and attack samples in third row is also more significant, indicating the good discriminative information with the results of STPM.  

Feature distribution of the testing videos on OULU-NPU Protocol 1 is shown in Fig.~\ref{fig:scatter_plot}. The right image (w/ RSGB) presents more well-clustered behavior than the left image (w/o RSGB), which demonstrates the excellent discrimination ability of our proposed RSGB for distinguishing the living and spoofing faces.

\section{Conclusions}

In this paper, we propose a novel face anti-spoofing method, which exploits fine-grained spatio-temporal information for facial depth estimation. In our method, Residual Spatial Gradient Block (RSGB) is utilized to detect more discriminative details while Spatio-Temporal Propagation Module (STPM) to encode spatio-temporal information. An extra Contrastive Depth Loss (CDL) is designed to improve the generality of depth-supervised PAD. 
We also investigate the effectiveness of actual depth map in face anti-spoofing. Extensive experiments demonstrate the superiority of our method. 

\section*{Acknowledgment}
This work has been partially supported by the Chinese National Natural Science Foundation Projects \#61876178, \#61806196, \#61976229.

\newpage

{\small
\bibliographystyle{ieee_fullname}
\bibliography{egbib}
}

\newpage

\section*{\Large Appendix}

\section{Temporal Depth in Face Anti-spoofing}
In this section, we use some simple examples to explain that exploiting temporal depth and motion is reasonable in the face anti-spoofing task.
\begin{figure}[htbp]
  \includegraphics[width=0.48\textwidth]{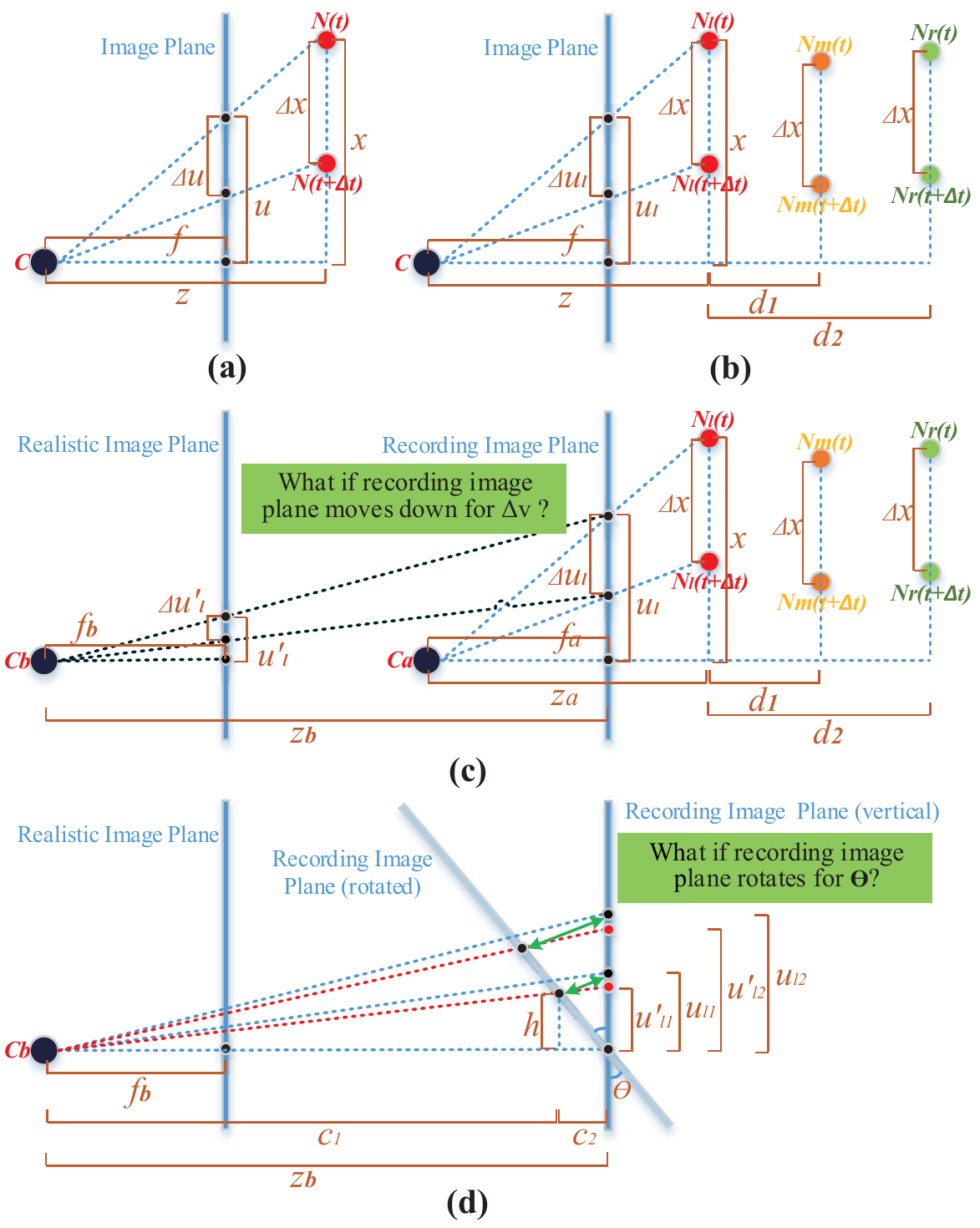}
  \caption{The schematic diagram of motion and depth variation in different scenes.}
  \label{fig:theory1}
\end{figure}

\subsection{Basic Scene}
As shown in Fig.~\ref{fig:theory1}(a), node $C$ denotes the camera focus. 
\textbf{Image Plane} represents the image plane of camera. $N(t)$ is one facial point at time $t$, and $N(t+\Delta t)$ is the corresponding point when $N(t)$ moves down vertically for $\Delta x$ at time $t+\Delta t$. For example, $N(t)$ can be the point of nose or ear. $f$ denotes the focal distance, and $z$ is the horizontal distance from the focal point to the point $N(t)$. $u$ and $x$ are the corresponding coordinates in vertical dimension. When $N(t)$ moves down vertically to $N(t+\Delta t)$ for $\Delta x$, the motion can be reflected on the image plane as $\Delta u$.
According to the camera model, we can obtain:
\begin{equation}
\begin{split}
& \frac{x}{u} = \frac{z}{f}, \\
\Leftrightarrow & u = \frac{fx}{z}.
\end{split}
\label{eq:camera_basis}
\end{equation}
When $N(t)$ moves down vertically for $\Delta x$ to $N(t+\Delta t)$, the $\Delta u$ can be achieved:
\begin{equation}
\begin{split}
\Delta u &= \frac{f \Delta x}{z}.
\end{split}
\label{eq:1}
\end{equation}

As shown in Fig.~\ref{fig:theory1}(b), to distinguish points $N_l$, $N_m$ and $N_r$, we transform Eq.~\ref{eq:1} and get $\Delta u_l$, $\Delta u_m$ and $\Delta u_r$ ($\Delta u_m$ and $\Delta u_r$ are not shown in the figure):
\begin{equation}
\begin{split}
\Delta u_l &= \frac{f \Delta x}{z},\\
\Delta u_m &= \frac{f \Delta x}{z+d_1}, \\
\Delta u_r &= \frac{f \Delta x}{z+d_2}, \\
\end{split}
\label{eq:2}
\end{equation}
where $d_1$ and $d_2$ are the corresponding depth difference.
From Eq.~\ref{eq:2}, there are:
\begin{equation}
\begin{split}
\frac{\Delta u_l}{\Delta u_m} = \frac{z+d_1}{z} = \frac{d_1}{z} + 1,\\
\frac{\Delta u_l}{\Delta u_r} = \frac{z+d_2}{z} = \frac{d_2}{z} + 1.
\end{split}
\label{eq:3}
\end{equation}
Removing $z$ from Eq.~\ref{eq:3}, $d_1/d_2$ can be obtained:
\begin{equation}
\begin{split}
\frac{d_1}{d_2} =
\displaystyle{  \frac{  \displaystyle{\frac{\Delta u_l}{\Delta u_m}} - 1}{  \displaystyle{\frac{\Delta u_l}{\Delta u_r}} - 1} },
\end{split}
\label{eq:4}
\end{equation}
In this equation, we can see that the relative depth $d_1/d_2$ can be estimated by the motion of three points, when $d_2 \neq 0$. The equations above are about the real scenes. In the following, we will introduce the derivation of attack scenes.

\subsection{Attack Scene}
\subsubsection{What if the attack carriers move?}
\label{sec:screen_shake}
As shown in Fig.~\ref{fig:theory1}(c), there are two image spaces in attack scenes: one is recording image space, where we replace $z, f$ by $z_a, f_a$, and the other is realistic image space, where we replace $z, f$ by $z_b, f_b$. In the recording image space, it's similar to Eq.~\ref{eq:2}:
\begin{equation}
\begin{split}
\Delta u_l &= \frac{f_a \Delta x}{z_a},\\
\Delta u_m &= \frac{f_a \Delta x}{z_a+d_1}, \\
\Delta u_r &= \frac{f_a \Delta x}{z_a+d_2}, \\
\end{split}
\label{eq:5}
\end{equation}
where $\Delta u_l, \Delta u_m, \Delta u_r$ are the magnitude of optical flow when three points $N_l(t), N_m(t), N_r(t)$ move down vertically for $\Delta x$.

In the realistic image space, there are:
\begin{equation}
\begin{split}
\Delta u'_l &= \frac{f_b \Delta x_l}{z_b},\\
\Delta u'_m &= \frac{f_b \Delta x_m}{z_b}, \\
\Delta u'_r &= \frac{f_b \Delta x_r}{z_b}, \\
\end{split}
\label{eq:6}
\end{equation}
where $\Delta x_l$, $\Delta x_m$ and $\Delta x_r$ are the motion of three points on the recording image plane, and $\Delta u_l, \Delta u_m, \Delta u_r$ are the corresponding values mapping on the realistic image plane.

Actually, there are $\Delta x_l = \Delta u_l, \Delta x_m= \Delta u_m, \Delta x_r= \Delta u_r$, if the recording screen is static. Now, a vertical motion $\Delta v$ is given to the recording screen, just as $\Delta x_l = \Delta u_l + \Delta v, \Delta x_m= \Delta u_m + \Delta v, \Delta x_r= \Delta u_r + \Delta v$. By inserting $\Delta v$, we transform Eq.~\ref{eq:6} into:
\begin{equation}
\begin{split}
\Delta u'_l &= \frac{f_a f_b \Delta x + z_a f_b \Delta v}{z_a z_b},\\
\Delta u'_m &= \frac{f_a f_b \Delta x + ( z_a + d_1 ) f_b \Delta v}{(z_a + d_1) z_b },\\
\Delta u'_r &= \frac{f_a f_b \Delta x + ( z_a + d_2 ) f_b \Delta v}{(z_a + d_2) z_b },\\
\end{split}
\label{eq:7}
\end{equation}
Due to that only $\Delta u'_l, \Delta u'_m, \Delta u'_r$ can be observed directly in the sequential images, we can estimate the relative depth via $\Delta u'_l, \Delta u'_m, \Delta u'_r$. So we leverage Eq.~\ref{eq:4} to estimate the relative depth $d'_1 / d'_2$:
\begin{equation}
\begin{split}
\frac{d'_1}{d'_2} =
\displaystyle{  \frac{  \displaystyle{\frac{\Delta u'_l}{\Delta u'_m}} - 1}{  \displaystyle{\frac{\Delta u'_l}{\Delta u'_r}} - 1} },
\end{split}
\label{eq:extra_1}
\end{equation}
and then we can insert Eq.~\ref{eq:7} into Eq.~\ref{eq:extra_1} to get:
\begin{equation}
\begin{split}
\frac{d'_1}{d'_2} = \frac{d_1}{d_2} \cdot \frac{f_a \Delta x + (z_a + d_2) \Delta v} {f_a \Delta x + (z_a + d_1) \Delta v}.
\end{split}
\label{eq:8}
\end{equation}
According to equations above, some important conclusions can be summarized:

\begin{itemize}
\item If $\Delta x = 0$, the scene can be recognized as print attack and Eq.~\ref{eq:8} will be invalid, for $\Delta u'_l = \Delta u'_r$, and the denominator in Eq.~\ref{eq:extra_1} will be zero. So here we use Eq.~\ref{eq:7} and
\begin{equation}
\begin{split}
\frac{\Delta u'_l}{\Delta u'_m} = \frac{d'_1}{z_b} + 1,\\
\frac{\Delta u'_l}{\Delta u'_r} = \frac{d'_2}{z_b} + 1,
\end{split}
\label{eq:extra_eq3}
\end{equation}
to obtain:
\begin{equation}
\begin{split}
d'_1 = d'_2 = 0.
\end{split}
\label{eq:12}
\end{equation}
In this case, it's obvious that the facial relative depth is abnormal and the face is fake. 
\item If $\Delta x \neq 0$, the scene can be recognized as replay attack.
\begin{itemize}
\item If $\Delta v = 0$, there is:
\begin{equation}
\begin{split}
\frac{d'_1}{d'_2} = \frac{d_1}{d_2}.
\end{split}
\label{eq:9}
\end{equation}
In this case, if these two image planes are parallel and the single-frame model can not detect the static spoof cues, the model will fail in the task of face anti-spoofing, owing to that the model is hard to find the abnormality of relative depth estimated from the facial motion. We call this scene \textbf{Perfect Spoofing Scene(PSS)}. Of course, making up \textbf{PSS} will cost a lot and is approximately impossible in practice.
\item If $\Delta v \neq 0$ and we want to meet Eq.~\ref{eq:9}, the following equation should be satisfied:
\begin{equation}
\begin{split}
\frac{f_a \Delta x + (z_a + d_2) \Delta v} {f_a \Delta x + (z_a + d_1) \Delta v} = 1,
\end{split}
\label{eq:10}
\end{equation}
then,
\begin{equation}
\begin{split}
& (d_2 - d_1) \Delta v  = 0, \\
\Leftrightarrow & d_2 - d_1 = 0, \ if \ \Delta v \neq 0.
\end{split}
\label{eq:10_2}
\end{equation}

However, in our assumption, $d_1 \neq d_2$, so:
\begin{equation}
\begin{split}
\frac{d'_1}{d'_2} \neq \frac{d_1}{d_2}.
\end{split}
\label{eq:11}
\end{equation}
This equation indicates that relative depth can't be estimated preciously, if the attack carrier moves in the replay attack. 
And $\Delta v$ usually varies when attack carrier moves in the long-term sequence, leading to the variation of $d'_1/d'_2$.
This kind of abnormality is more obvious along with the long-term motion.
\end{itemize}

\item If $d_2$ denotes the largest depth difference among facial points, then $d_1/d_2 \in [0, 1]$, showing that constraining depth label of living face to $[0, 1]$ is valid. As analyzed above, for spoofing scenes, the abnormal relative depth usually varies over time, so it is too complex to be computed directly. Therefore, we merely set depth label of spoofing face to all 0 to distinguish it from living label, making the model learn the abnormity under depth supervision itself.

\end{itemize}

\subsubsection{What if the attack carriers rotate?}
As shown in Fig.~\ref{fig:theory1}(d), we rotate the recording image plane for degree $\theta$. $u_{l2}, u_{l1}$ are the coordinates of $N_l(t), N_l(t+\Delta t)$ mapping on the recording image plane.
The two black points at the \emph{right} end of green double arrows on recording image plane (vertical) will reach the two black points at the \emph{left} end of green double arrow on recording image plane (rotated), when the recording image plane rotates. And the corresponding values $u_{l2}, u_{l1}$ will not change after rotation.
For convenient computation, we still map the rotated points to the vertical recording image plane. And the coordinates after mapping are $u'_{l2}, u'_{l1}$. $c_1, c_2, h$ are the corresponding distances shown in the figure. According to the relationship of the  foundamental variables, we can obtain:
\begin{equation}
\begin{split}
& h = u_{l1} \cos{\theta}, \\
& \frac{z_b}{c_{1}} = \frac{u'_{l1}}{h}, \\
& c_2 = u_{l1} \sin{\theta}, \\
& c_{1} + c_{2} = z_b.
\end{split}
\label{eq:rotation_1}
\end{equation}
Deriving from equations above, we can get $u'_{l1}$:
\begin{equation}
\begin{split}
u'_{l1} = \frac{z_b u_{l1} \cos{\theta}}{z_b - u_{l1} \sin{\theta}},
\end{split}
\label{eq:rotation_2}
\end{equation}
and $u'_{l2}$ can also be calculated by imitating Eq.~\ref{eq:rotation_2}:
\begin{equation}
\begin{split}
u'_{l2} = \frac{z_b u_{l2} \cos{\theta}}{z_b - u_{l2} \sin{\theta}}.
\end{split}
\label{eq:rotation_3}
\end{equation}
Subtract $u'_{l1}$ from $u'_{l2}$, the following is achieved:
\begin{equation}
\begin{split}
u'_{l2} - u'_{l1} = (u_{l2} - u_{l1}) \cdot \frac{{z_b}^2 \cos{\theta}}{ (z_b - u_{l1} \sin{\theta}) (z_b - u_{l2} \sin{\theta})}.
\end{split}
\label{eq:rotation_4}
\end{equation}
Obviously, $u_{l2} - u_{l1} = \Delta u_l$. We define $u'_{l2} - u'_{l1} = \Delta u^{\theta}_l$. And then we get the following equation:
\begin{equation}
\begin{split}
\Delta u^{\theta}_l &= \Delta u_l \cdot \frac{{z_b}^2 \cos{\theta}}{ (z_b - u_{l1} \sin{\theta}) [z_b - (u_{l1}+\Delta u_l) \sin{\theta}]}, \\
\Delta u^{\theta}_m &= \Delta u_m \cdot \frac{{z_b}^2 \cos{\theta}}{ (z_b - u_{m1} \sin{\theta}) [z_b - (u_{m1}+\Delta u_l) \sin{\theta}]}, \\
\Delta u^{\theta}_r &= \Delta u_r \cdot \frac{{z_b}^2 \cos{\theta}}{ (z_b - u_{r1} \sin{\theta}) [z_b - (u_{r1}+\Delta u_l) \sin{\theta}]},
\end{split}
\label{eq:rotation_5}
\end{equation}
where the relationship between $\Delta u^{\theta}_m, \Delta u^{\theta}_r$ and $N_m(t), N_r(t)$ are just like that between $\Delta u^{\theta}_l$ and $N_l(t)$, as well as $u_{m1}, u_{r1}$. Note that for simplification, we only discuss the situation that $u_{l1}, u_{m1}, u_{r1}$ are all positive.

Reviewing Eq.~\ref{eq:6}, We can confirm that $\Delta x_l = \Delta u^{\theta}_l$, $\Delta x_m = \Delta u^{\theta}_m$, $\Delta x_r = \Delta u^{\theta}_r$. According to Eq.~\ref{eq:extra_1}, the final $d'_1/d'_2$ can be estimated:
\begin{equation}
\begin{split}
\frac{d'_1}{d'_2} =
\displaystyle{  \frac{  \displaystyle{\frac{\Delta u_l}{\Delta u_m}} \cdot \beta_1 - 1}{  \displaystyle{\frac{\Delta u_l}{\Delta u_r}} \cdot \beta_2 - 1} } =
\displaystyle{  \frac{  (\displaystyle{\frac{d_1}{z_a}}+1) \cdot \beta_1 - 1}{  (\displaystyle{\frac{d_2}{z_a}}+1)  \cdot \beta_2 - 1} },
\end{split}
\label{eq:rotation_6}
\end{equation}
where $\beta_1$ and $\beta_2$ can be represented as:
\begin{equation}
\begin{split}
\beta_1 &= \frac{ (z_b - u_{m1}\sin{\theta})(z_b-u_{m2}\sin{\theta}) }{ (z_b - u_{l1}\sin{\theta})(z_b-u_{l2}\sin{\theta}) }, \\
\beta_2 &= \frac{ (z_b - u_{r1}\sin{\theta})(z_b-u_{r2}\sin{\theta}) }{ (z_b - u_{l1}\sin{\theta})(z_b-u_{l2}\sin{\theta}) },
\end{split}
\label{eq:rotation_7}
\end{equation}
where $u_{l2} = u_{l1} + \Delta u_l, u_{m2} = u_{m1} + \Delta u_m, u_{r2} = u_{r1} + \Delta u_r$. Observing Eq.~\ref{eq:rotation_6}, we can see if $\beta_1 < 1, \beta_2 > 1$ or $\beta_1 >1, \beta_2 < 1$, there will be $d'_1/d'_2 \neq d_1/d_2$, .

Now, we discuss the sufficient condition of $\beta_1 < 1, \beta_2 > 1$. When $u_{m1} > u_{l1}, u_{m2} > u_{l2}, u_{r1} < u_{l1}, u_{r2} < u_{l2}$, the $\beta_1 < 1, \beta_2 > 1$ can be established. Similar to Eq.~\ref{eq:2}, the relationship of variables can be achieved:
\begin{equation}
\begin{split}
& \frac{f_a x_{l1}}{z_a} = u_{l1}, \frac{f_a x_{l2}}{z_a} = u_{l2}, \\
& \frac{f_a x_{m1}}{z_a + d_1} = u_{m1}, \frac{f_a x_{m2}}{z_a + d_1} = u_{m2}, \\
& \frac{f_a x_{r1}}{z_a + d_2} = u_{r1}, \frac{f_a x_{r2}}{z_a + d_2} = u_{r2},
\end{split}
\label{eq:rotation_extra_relation}
\end{equation}
From Eq.~\ref{eq:rotation_extra_relation} and $u_{m1} > u_{l1}, u_{m2} > u_{l2}$, we can obtain:
\begin{equation}
\begin{split}
x_{m1} &> x_{l1} \cdot \frac{z_a+d_1}{z_a}, \\
x_{m1} + \Delta x &> (x_{l1}+\Delta x) \cdot \frac{z_a+d_1}{z_a}, \\
\end{split}
\label{eq:rotation_8}
\end{equation}
\begin{equation}
\begin{split}
x_{r1} &< x_{l1} \cdot \frac{z_a+d_2}{z_a}, \\
x_{r1} + \Delta x &< (x_{l1}+\Delta x) \cdot \frac{z_a+d_2}{z_a}, \\
\end{split}
\label{eq:rotation_9}
\end{equation}
where $x_{l1}, x_{l2}, x_{m1}, x_{m2}, x_{r1}, x_{r2}$ are corresponding coordinates of $N_l(t+\Delta t), N_l(t), N_m(t+\Delta t), N_m(t), N_r(t+\Delta t), N_r(t)$ in the dimension of $\textbf{x}$ in recording image space. In facial regions, we can easily find corresponding points $N_l(t), N_m(t)$, which satisfy that $d_1 \ll z_a$ (i.e., $d_1 = 0$) and $x_{m1} > x_{l1}$. In this pattern, Eq.~\ref{eq:rotation_8} can be established. To establish Eq.~\ref{eq:rotation_9}, we only need to find point $N_r(t)$, which satisfies that $x_{r1} < x_{l1}$. According to the derivation above, we can see that there exists cases that $d'_1/d'_2 < d_1/d_2$. And there are also many cases that satisfy $d'_1/d'_2 > d_1/d_2$, which we do not elaborate here. When faces move, the absolute coordinates $x_{l1}, x_{l2}, x_{m1}, x_{m2}, x_{r1}, x_{r2}$ vary, as well as $\beta_1, \beta_2$, leading to the variation of estimated relative depth of three facial points at different moments, which will not occur in the \emph{real} scene. That's to say, if the realistic image plane and recording image plane are not parallel, we can seek cases to detect abnormal relative depth with the help of abnormal facial motion.

\subsubsection{Discussion}
One of basis of the elaboration above is that the structure of face is similar to that of the hill, which is complex, dense and undulate. This is interesting and worth being exploited in face anti-spoofing.

Even though we only use some special examples to demonstrate our viewpoints and assumption, they can still prove the reasonability of utilizing facial motion to estimate the relative facial depth in face anti-spoofing task. In this way, the learned model can seek the abnormal relative depth and motion in the facial regions. And our extensive experiment demonstrates our assumption and indicates that temporal depth method indeed improves the performance of face anti-spoofing.

\end{document}